\def\year{2021}\relax
\documentclass[letterpaper]{article} 
\usepackage{aaai21} 
\usepackage{times} 
\usepackage{helvet} 
\usepackage{courier} 
\usepackage[hyphens]{url} 
\usepackage{graphicx} 
\usepackage{natbib} 
\usepackage{caption} 
\usepackage{multirow}
\usepackage{algorithm}
\usepackage{algorithmic}
\usepackage{amsfonts}
\usepackage{amsmath}
\usepackage{subfigure}
\usepackage{color}
\usepackage{url}
\usepackage[switch]{lineno}
\title{Local Relation Learning for Face Forgery Detection}
\author {
    Shen Chen\textsuperscript{\rm 1,2}\footnote{Equal contribution.},
    Taiping Yao\textsuperscript{\rm 2}\footnotemark[\value{footnote}],
    Yang Chen\textsuperscript{\rm 2},
    Shouhong Ding\textsuperscript{\rm 2}\footnote{Corresponding authors.},
    Jilin Li\textsuperscript{\rm 2},
    Rongrong Ji\textsuperscript{\rm 1,3}\footnotemark[\value{footnote}] \\
}
\affiliations {
    \textsuperscript{\rm 1} Media Analytics and Computing Lab, Department of Artificial Intelligence, \\ School of Informatics, Xiamen University \\
    \textsuperscript{\rm 2} YouTu Lab, Tencent \\
    \textsuperscript{\rm 3} Institute of Artificial Intelligence, Xiamen University \\
    chenshen@stu.xmu.edu.cn, rrji@xmu.edu.cn, \\
    \{taipingyao, wizyangchen, ericshding, jerolinli\}@tencent.com
}

\begin{document} 

\maketitle

\begin{abstract}
With the rapid development of facial manipulation techniques, face forgery detection has received considerable attention in digital media forensics due to security concerns.
Most existing methods formulate face forgery detection as a classification problem and utilize binary labels or manipulated region masks as supervision.
However, without considering the correlation between local regions, these global supervisions are insufficient to learn a generalized feature and prone to overfitting.
To address this issue, we propose a novel perspective of face forgery detection via local relation learning.
Specifically, we propose a Multi-scale Patch Similarity Module (MPSM), which measures the similarity between features of local regions and forms a robust and generalized similarity pattern.
Moreover, we propose an RGB-Frequency Attention Module (RFAM) to fuse information in both RGB and frequency domains for more comprehensive local feature representation, which further improves the reliability of the similarity pattern. 
Extensive experiments show that the proposed method consistently outperforms the state-of-the-arts on widely-used benchmarks.
Furthermore, detailed visualization shows the robustness and interpretability of our method.
\end{abstract}

\section{Introduction}
Recent studies have shown rapid progress in facial manipulation, which enables an attacker to manipulate or forge the facial area of human faces, such as Deepfakes~\cite{Tora2018DF} and FaceSwap~\cite{Kowalski2018FS}.
With the remarkable success in synthesizing realistic faces, it becomes infeasible even for humans to distinguish whether an image has been manipulated.
At the same time, these forged images may be abused for malicious purposes, causing severe trust issues and security concerns in our society.
Therefore, it is of paramount importance to develop effective methods for detecting face forgery.

Early works~\cite{Afchar2018MesoNetAC,Nguyen2019CapsuleforensicsUC,Rssler2019FaceForensicsLT} treat this challenge as a binary classification problem and develop CNNs to model the decision boundary between real and forged faces.
However, this setting is known to easy to overfit and lacks interpretability.
To relieve this problem, recent works~\cite{Nguyen2019MultitaskLF,Stehouwer2020OnTD} introduce the manipulated region mask as supervision to assist networks in locating specific forged regions.
Although these methods have achieved remarkable performance on high-quality images, they are vulnerable to disturbances like image compression or noises.
Moreover, the generalization issue is still not well addressed. When applied to forgery generated by unseen face manipulation methods, they experience a significant performance drop.
Face X-ray~\cite{Li2020FaceXF} has noticed this generalization problem and designs new supervision named face X-ray, which focuses on the artifacts caused by image blending between two images.
However, it can not be applied to forgery methods without using blending and also be greatly affected by image noises.
This means these methods have not learned the intrinsic characteristics in face forgery detection.
In a word, most current methods use a kind of global supervision by utilizing binary labels or manipulated region masks at the end of the network, but still fail to learn robust and generalized features.

\begin{figure*}[t]
	\begin{center}
		\includegraphics[width=0.95\textwidth]{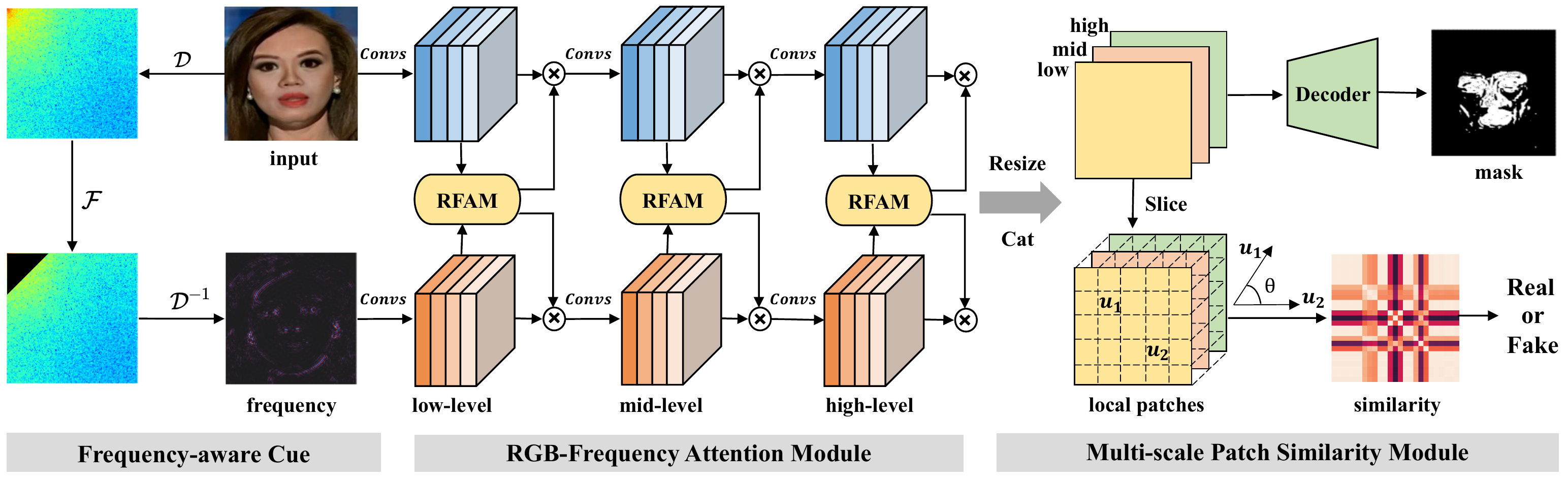}
	\end{center}
	\caption{
 		The framework of the proposed Local Relation Learning for Face Forgery Detection.
	}
	\label{framework}
\end{figure*}

To address the above issues, we propose a novel perspective of face forgery detection via local relation learning to focuses on the relation of local regions, which is a generalized descriptor that can be used to effectively capture the forged trace, such as abnormal texture~\cite{Liu2020GlobalTE} and high frequency noise~\cite{Qian2020ThinkingIF}. 
Specifically, we design a Multi-scale Patch Similarity Module (MPSM) to explicitly model a second-order relationship between different local regions and build a similarity pattern with pair-wise cosine measurement. In this similarity pattern, real regions are similar to each other, so as the forged regions. But real regions and forged regions are not similar. By encoding local forgery semantics which is more robust against disturbances and less sensitive to specific manipulation methods, the similarity pattern further connects them and builds a global face forgery representation, which is more robust and 
generalized for forgery detection.

Moreover, to further improve the reliability of the similarity pattern, we propose an RGB-Frequency Attention Module (RFAM) to both consider RGB and frequency information and collaboratively learn a comprehensive local representation.
Concretely, we first transform the image into frequency domain and amplify the artifacts hidden in high frequency based on \textit{Discrete Cosine Transform} (DCT), then a two-stream network with attention modules in different blocks is designed to fuse RGB and frequency features, improving the integrity of local representation.

The main contributions of this work are summarized as follows:

\begin{itemize}
    \item We address face forgery detection via local relation learning, and propose a similarity pattern which is a generalized descriptor that can be used to effectively capture the forged trace.
    \item We combine RGB and frequency information based on the attention module to collaboratively learn comprehensive representation, which further improves the reliability of the similarity pattern.
    \item Extensive experiments and visualizations are presented to reveal the robustness and generalization of the proposed similarity pattern, which demonstrates the effectiveness of our method against the state-of-the-art competitors.
\end{itemize}

\section{Related Works}
Over the past several years, forgery creation has recently gained significant attention.
With the complementary property of forgery creation and detection, face forgery detection also becomes an increasingly emerging research area.
In this section, we briefly review prior image forgery methods including face forgery to which our method belongs.

Face forgery detection is mostly regarded as merely a binary (real or forgery) classification problem.
With the tremendous success of deep learning~\cite{Chollet2017XceptionDL,Tan2019EfficientNetRM,Lin2020HRankFP}, some works~\cite{Afchar2018MesoNetAC,Nguyen2019CapsuleforensicsUC,Rssler2019FaceForensicsLT} adopt neural networks to automatically extract discriminative features for forgery detection.  However, they are easy to overfit and lacks interpretability.
Besides classification, there are methods focusing on localizing the manipulated region.
Some works~\cite{Bappy2017ExploitingSS,Nguyen2019MultitaskLF,Salloum2018ImageSL} use multi-task learning to simultaneously classify the manipulated images and locate the manipulated region.
Instead of simply using a multi-task learning strategy, Stehouwer \emph{et al.}~\cite{Stehouwer2020OnTD} highlights the informative regions through an attention mechanism where the attention map is guided by the ground truth manipulation mask. 
Face X-ray~\cite{Li2020FaceXF} localizes the forged boundary in a self-supervised mechanism, achieving remarkable performance at high-quality data. Although these works have achieved considerable performance, they are easily affected by image compression or noise, leading to poor generalizability.

Recently, several attempts have been made to solve forgery detection using frequency cues.
For example, Durall \emph{et al.}~\cite{Durall2019UnmaskingDW} extracts frequency-domain information using DFT transform and averages the amplitudes of different frequency bands.
Two-branch RN~\cite{Masi2020TwobranchRN} amplifies multi-band frequencies using a Laplacian of Gaussian (LoG) as a bottleneck layer.
F3 Net~\cite{Qian2020ThinkingIF} takes advantage of frequency-aware decomposed image components and local frequency statistics to deeply mine the forged patterns. Although these methods extract features more comprehensively, they still cannot guarantee strong generalization. To learn more robust and generalized features, we address face forgery detection via local relation learning, which focuses on the relation of local regions.

\section{Approach}
The framework of the proposed approach is illustrated in Fig.\,\ref{framework}.
We will present the extraction strategy for frequency information in Sec.\,\ref{fac}, and the proposed RFAM for co-learning RGB and frequency domain information in Sec.\ref{rfam}.
We also describe the MPSM for learning local region associations in Sec.\,\ref{mspsm}.
All the supervised loss functions are summarized in Sec.\,\ref{loss_funcdtions}.

\subsection{Frequency-aware Cue} \label{fac}
As stated in~\cite{Zhang2019DetectingAS}, most existing face manipulated methods are based on GAN, where the up-sampling operation causes anomalies in the frequency statistical properties of the forged faces.
To this end, we introduce frequency information to assist the network in mining the essential difference between real and forged regions.

Without loss of generality, let $\mathbf{x}_1 \in \mathbb{R}^{H \times W \times 3}$ denote the input RGB image, where $H$ and $W$ denote the height and width of image, respectively.
We first transform $\mathbf{x}_{1}$ from RGB domain to frequency domain. That is:
\begin{equation}
 \mathbf{x}_{1}^{d} = \mathcal{D}(\mathbf{x}_1),
\end{equation}
where $\mathbf{x}_{1}^{d} \in \mathbb{R}^{H \times W \times 1}$, and $\mathcal{D}$ denotes the \textit{Discrete Cosine Transform} (DCT), according to its wide applications in image processing, and its nice layout of the frequency distribution, \emph{i.e.}, low frequency responses are placed in the top-left corner, and high frequency responses are located in the bottom-right corner.

We then suppress the image content by filtering out the low frequency information to amplify subtle artifacts at high frequencies.
That is:
\begin{equation}
 \mathbf{x}_{1}^{f} = \mathcal{F}(\mathbf{x}_{1}^{d}, \alpha),
\end{equation}
where $\mathcal{F}$ denotes high pass filtering and $\alpha$ controls the low frequency component to be filtered out.
Specifically, $\mathcal{F}$ sets the triangle with length $\alpha$ on the top-left corner of $\mathbf{x}_{1}^{d}$ to 0.

Since the frequency domain does not match the shift-invariance and local consistency owned by natural images, we invert $\mathbf{x}_{1}^{f}$ back into RGB color space via $\mathcal{D}^{-1}$ to obtain the desired representation at frequency domain.
The entire process can be summarized as follows:
\begin{equation}
 \label{freq}
 \mathbf{x}_{2} = \mathcal{D}^{-1}(\mathcal{F}(\mathcal{D}(\mathbf{x}_1), \alpha)),
\end{equation}
where $\mathbf{x}_{2} \in \mathbb{R}^{H \times W \times 1}$.
We will explain in detail how frequency domain information aids attention to subtle artifacts in the visualization sec.\,\ref{visulization}.

\begin{figure}[t]
	\centering
	\includegraphics[width=0.45\textwidth]{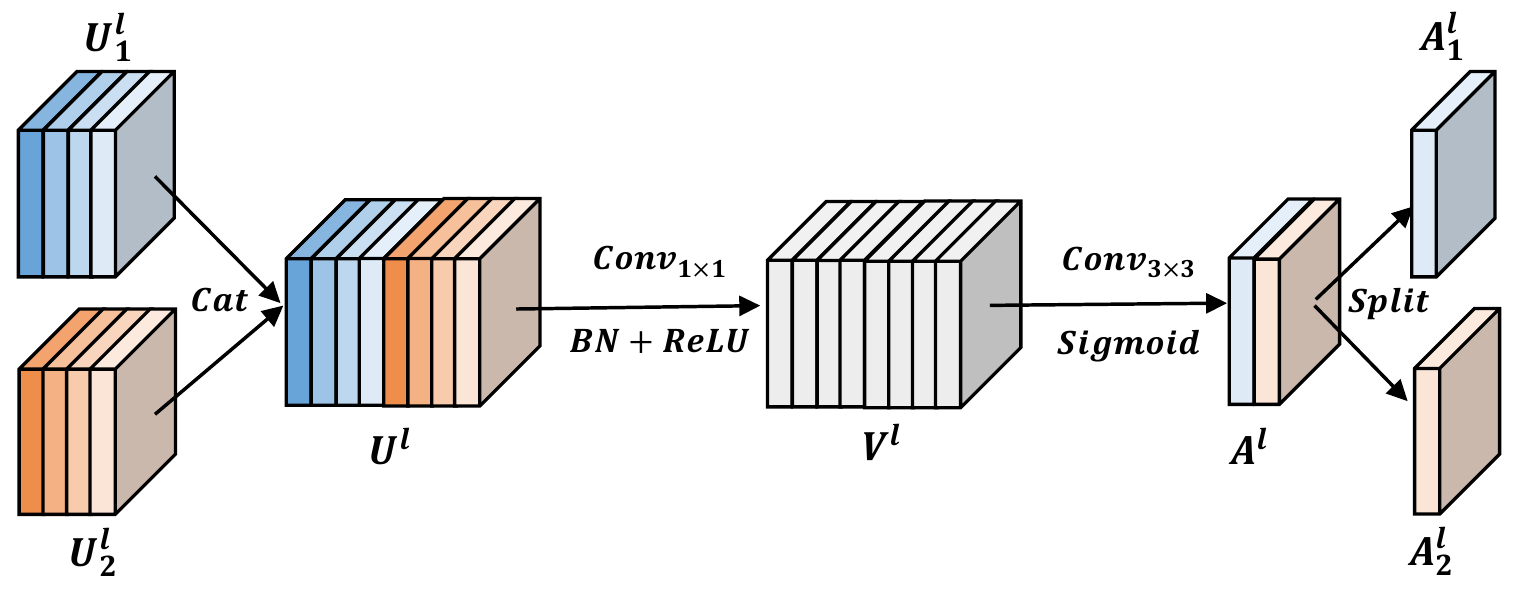}
	\caption{RGB-Frequency Attention Module.}
	\label{fig_rfam}
\end{figure}

\subsection{RGB-Frequency Attention Module} \label{rfam}
In this section, we develop a two-stream network whose input contains RGB data $\mathbf{x}_1$ and frequency data $\mathbf{x}_{2}$.
For forged images, RGB information is useful for locating anomalous textures, while frequency information amplifies the subtle manipulated artifacts.
Fusing them contributes to a more comprehensive feature representation. 
To take full advantage of this information, we design an RGB Frequency Attention Module (RFAM) that collaboratively fuse RGB and frequency information at different semantic layers, facilitating the learning of local region features.

As shown in Fig.\,\ref{fig_rfam}, $\mathbf{U}_1^l \in \mathbb{R}^{H^l \times W^l \times C^l}$ and $\mathbf{U}_2^l \in \mathbb{R}^{H^l \times W^l \times C^l}$ represent the feature map of the frequency stream and the RGB stream at the $l$-th layer of network, respectively.
To simplify, we let $l \in \{low, mid, high\}$ denote the different semantic layers of the network, and $H^l$, $W^l$ , and $C^l$ be the height, width, and channel of the feature map of the corresponding layer.

We first concatenate $\mathbf{U}_1^l$ and $\mathbf{U}_2^l$ in the channel dimension to get $\mathbf{U}^l \in \mathbb{R}^{H \times W \times 2C}$.
Then $\mathbf{U}^l$ is through a $1 \times 1 $ convolution with BatchNormalization~\cite{Ioffe2015BatchNA} and ReLU non-linearity to comprehensively fuse the RGB and frequency information, which is formulated as:
\begin{equation}
 \mathbf{V}^l = \delta(\mathcal{B}(\textit{Conv}_{1 \times 1}(\mathbf{U}^l))),
\end{equation}
where $\mathbf{V}^l \in \mathbb{R}^{H \times W \times 2C}$, $\mathcal{B}$ denote BatchNormalization and $\delta$ is the ReLU function.

Furthermore, an attention map $\mathbf{A}^{l} \in \mathbb{R}^{H \times W \times 2}$ is created to highlight the regions of interest in $\mathbf{U}_1^l$ and $\mathbf{U}_2^l$, respectively. That is:
\begin{equation}
 \mathbf{A}^l = \sigma(\textit{Conv}_{3 \times 3}(\mathbf{V}^l)),
\end{equation}
where $\textit{Conv}_{3 \times 3}$ is a $3 \times 3$ convolution, $\sigma$ denotes the \textit{Sigmoid} function.
Finally, we split $\mathbf{A}^{l}$ into $\mathbf{A}_{1}^{l} \in \mathbb{R}^{H^l \times W^l \times 1}$ and ${A}_{2}^{l} \in \mathbb{R}^{H^l \times W^l \times 1}$, which correspond to the attention map of RGB stream and frequency stream, respectively.
The most significant spatial regions are highlighted via dot products between the feature map and attention map.

The proposed RFAM collaboratively fuses the forged cues under the RGB and frequency domains, facilitating the learning of the correlation between local regions.

\begin{table*}[t]
\centering
\begin{tabular}{l|cccccc}
\hline
\multirow{2}{*}{\textbf{Methods}} & \textbf{ACC} & \textbf{AUC} & \textbf{ACC} & \textbf{AUC} & \textbf{ACC} & \textbf{AUC}    \\ \cline{2-7} & \textbf{(Raw)} & \textbf{(Raw)} & \textbf{(HQ)} & \textbf{(HQ)} & \textbf{(LQ)} & \textbf{(LQ)} \\ \hline
Steg.Features~\cite{Fridrich2012RichMF} & 97.63\% & - & 70.97\% & - & 55.98\% & - \\ \hline
LD-CNN~\cite{Cozzolino2017RecastingRL} & 98.57\% & - & 78.45\% & - & 58.69\% & - \\ \hline
C-Conv~\cite{Bayar2016ADL} & 98.74\% & - & 82.97\%   & -    & 66.84\%   & -    \\ \hline
CP-CNN~\cite{Rahmouni2017DistinguishingCG} & 97.03\% & -  & 79.08\%   & -    & 61.18\%   & -    \\ \hline
MesoNet~\cite{Afchar2018MesoNetAC} & 95.23\% & -  & 83.10\%   & -    & 70.47\%   & -    \\ \hline
Xception~\cite{Rssler2019FaceForensicsLT} & 99.26\% & -  & 95.73\%   & -    & 86.86\%   & -    \\ \hline
Face X-ray~\cite{Li2020FaceXF} & -  & -  & -    & 87.40\%   & -    & 61.60\%   \\ \hline
Two-branch RN~\cite{Masi2020TwobranchRN} & -  & -  & 96.43\%   & 88.70\%   & 86.34\%   & 86.59\%   \\ \hline
F3-Net~\cite{Qian2020ThinkingIF} & \textbf{99.95\%} & 99.80\% & 97.52\%   & 98.10\%   & 90.43\%   & 93.30\%   \\ \hline
Ours & 99.87\% & \textbf{99.92\%} & \textbf{97.59\%} & \textbf{99.46\%} & \textbf{91.47\%} & \textbf{95.21\%} \\ \hline
\end{tabular}
\caption{Quantitative results in terms of ACC and AUC on FaceForensics++ dataset with all quality settings, \emph{i.e.}, low quality (LQ), high quality (HQ), and raw videos without compression (Raw).}
\label{faceforensics}
\end{table*}

\subsection{Multi-scale Patch Similarity Module} \label{mspsm}
Unlike existing methods that discriminate based on the global pooling of feature map, we propose a Multi-scale Patch Similarity Module (MPSM), which measures the similarity of local regions based on the learned RGB and frequency features.

Specifically, we first fuse the results from $l$-th layer of the two-stream network, as follows:
\begin{equation}
 \mathbf{\widetilde{U}}^l = \mathbf{A}_1^{l} \odot \mathbf{U}_1^{l} + \mathbf{A}_2^{l} \odot \mathbf{U}_2^{l},
\end{equation}
where $\mathbf{\widetilde{U}}^l \in \mathbb{R}^{H^l \times W^l \times C^l}$ denotes the fused feature map of $l$-th layer.
Through summing over the spatial dimension, we can directly merge the regions of interest in the RGB and frequency domains.

To address the problem of different forged region sizes, we introduce the multi-scale features, where high-resolution low-level features facilitate localization and high-level features rich in semantic information are used for identification.
Specifically, we resize $\mathbf{\widetilde{U}}^{low}$ and $\mathbf{\widetilde{U}}^{mid}$ to the same size as $\mathbf{\widetilde{U}}^{high}$, and then concatenate them together to get the multi-scale features $\mathbf{\widetilde{U}} \in \mathbb{R}^{\widetilde{H} \times \widetilde{W} \times \widetilde{C}}$, where $\widetilde{H} = H^{high}$, $\widetilde{W} = W^{high}$, and $\widetilde{C} = \sum_{l \in {\{low, mid, high\}} } C^l$.

To construct associations between local regions, we partition $\mathbf{\widetilde{U}}$ spatially into $k \times k$ patches $\mathbf{\widetilde{u}}_i \in \mathbb{R}^{h \times w \times \widetilde{C}}$, where $i \in \{1,2,\dots,k^2\}$, $h = \textit{Ceil}(\frac{\widetilde{H}}{k})$ and $w = \textit{Ceil}(\frac{\widetilde{W}}{k})$.
Then $\mathbf{\widetilde{u}}_i$ is flattened into a one-dimensional vector $\mathbf{u}_i \in \mathbb{R}^{hw\widetilde{C}}$, and the similarity between patch $u_{i}$ and $u_{j}$ is calculated based on cosine distance.
That is:
\begin{equation}
\hat{\mathbf{s}}_{i, j}=\frac{\left\langle\frac{u_{i}}{\left\|u_{i}\right\|_{2}}, \frac{u_{j}}{\left\|u_{j}\right\|_{2}}\right\rangle+1}{2},
\end{equation}
where $\hat{\mathbf{s}}_{i j}$ ranges from 0 and 1, and $\hat{\mathbf{s}} \in \mathbb{R}^{k^2 \times k^2}$.
A lower $\hat{\mathbf{s}}_{i j}$ indicates a larger discrepancy between patches, \emph{i.e.}, one is the real and the other is forged, and vice versa.

We introduce the manipulated region mask and transform it into a kind of second-order supervision to guide the learning of similarity $\hat{\mathbf{s}}$.
To construct the mask, we pair forged images with their corresponding source images, compute the absolute pixel-wise difference in the RGB channels, convert into grayscale, and then divide by 255 to produce a map in the range of [0, 1].
We empirically determine the threshold of 0.15 to obtain the desired binary mask $\mathbf{M}$.

Without loss of generality, let $\mathbf{M} \in \mathbb{R}^{H \times W}$ denote the mask, where the value of the real region is 0 and the value of the forged region is 1.
That is:
\begin{equation}
\mathbf{M}_{i j}=\left\{\begin{array}{ll}
0 & \text { if } \mathbf{x}_{i j} \text { is real } \\
1 & \text { if } \mathbf{x}_{i j} \text { is forged }
\end{array}\right..
\end{equation}

Subsequently, we divide $\mathbf{M}$ into $k \times k$ patches $\mathbf{m}_i \in \mathbb{R}^{h \times w}$, where $i \in \{1,2,\dots,k^2\}$, $h = \textit{Ceil}(\frac{H}{k})$ and $w = \textit{Ceil}(\frac{W}{k})$.
Then the forged probability $\mathbf{p}_i \in [0, 1]$ for each patch $\mathbf{m}_i$ is obtained by averaging all the pixel values of $\mathbf{m}_i$.
Finally, based on the Euclidean distance between the forged probabilities, we obtain the desired second-order supervision, \emph{i.e.}, the relation of local regions.
That is:
\begin{equation}
\mathbf{s}_{i j}=1-\left(\mathbf{p}_{i}-\mathbf{p}_{j}\right)^{2},
\end{equation}
where $\mathbf{s}_{i j}$ ranges from 0 and 1, and $\mathbf{s} \in \mathbb{R}^{k^2 \times k^2}$ guides the learning of local relation.
Formally, we formulate the \textbf{Similarity Loss} function as follows:
\begin{equation}
\label{loss_sim}
\mathcal{L}_{sim}=\|\mathbf{s}-\hat{\mathbf{s}}\|_{2}.
\end{equation}

\begin{table}[t]
\centering
\setlength{\tabcolsep}{1.8mm}
\begin{tabular}{l|c|c|c|c}
\hline
\textbf{Methods} & \textbf{DF}  & \textbf{FF}  & \textbf{FS}  & \textbf{NT}  \\ \hline
Steg.Features  & 67.00\%   & 48.00\%   & 49.00\%   & 56.00\%   \\ \hline
LD-CNN   & 75.00\%   & 56.00\%   & 51.00\%   & 62.00\%   \\ \hline
C-Conv & 87.00\%   & 82.00\%   & 74.00\%   & 74.00\%   \\ \hline
CP-CNN & 80.00\%   & 62.00\%   & 59.00\%   & 59.00\%   \\ \hline
MesoNet   & 90.00\%   & 83.00\%   & 83.00\%   & 75.00\%   \\ \hline
Xception   & 96.01\%   & 93.29\%   & 94.71\%   & 79.14\%   \\ \hline
F3-Net   & 97.97\%   & 95.32\%   & 96.53\%   & 83.32\%   \\ \hline
Ours    & \textbf{98.84\%} & \textbf{95.53\%} & \textbf{97.53\%} & \textbf{89.31\%} \\ \hline
\end{tabular}
\caption{Quantitative results (ACC) on FaceForensics++ (LQ) with four manipulation methods, \emph{i.e.} DeepFakes (DF), Face2Face (FF), FaceSwap (FS) and NeuralTextures (NT).}
\label{single_test}
\end{table}

\begin{table*}[t]
\centering
\begin{tabular}{l|c|c|c|c|c|c|c}
\hline
\multirow{3}{*}{\textbf{Methods}} & \multirow{3}{*}{\textbf{Training dataset}} & \multicolumn{6}{c}{\textbf{Test datset}} \\ \cline{3-8} &
& \multicolumn{2}{c|}{\textbf{Celeb-DF}} & \multicolumn{2}{c|}{\textbf{DFDC}} & \multicolumn{2}{c}{\textbf{DFD}} \\ \cline{3-8} 
& & AUC & EER & AUC & EER & AUC & EER \\ \hline
Xception & FF++ & 36.19 & 59.64 & 48.98 & 50.45 & 87.86 & 21.49 \\ \hline
Ours & FF++ & \textbf{78.26} & \textbf{29.67} & \textbf{76.53} & \textbf{32.41} & \textbf{89.24} & \textbf{20.32} \\ \hline
\end{tabular}
\caption{Benchmark results in terms of AUC and EER for our framework and Xception~\cite{Rssler2019FaceForensicsLT} on unseen datasets.}
\label{cross_dataset}
\end{table*}

\subsection{Loss Functions} \label{loss_funcdtions}
We have learned a second-order pattern based on the association of local regions, which is used to identify whether a face is forged or not.
To this end, we flatten $\hat{\mathbf{s}}$ and successively pass it through the fully connected layer and \textit{Sigmoid} function to obtain the final predicted probability $\hat{\mathbf{y}}$.
And the \textbf{Cross-Entropy Loss} function is defined as:
\begin{equation}
\label{loss_ce}
\mathcal{L}_{\mathrm{ce}}=-\left[\mathbf{y} \log \hat{\mathbf{y}}+(1-\mathbf{y}) \log (1-\hat{\mathbf{y}})\right],
\end{equation}
where $\mathbf{y}$ is set to 1 if the face image has been manipulated, otherwise it is set to 0.

To compensate for the loss of edge information due to averaging operations in $\mathbf{m}_i$, we further introduce a decoder module to locate the specific forged region $\hat{\mathbf{M}} \in \mathbb{R}^{H \times W}$, and formulate the \textbf{Segmentation Loss} function as:
\begin{equation}
\label{loss_seg}
\mathcal{L}_{\mathrm{seg}}=\sum_{i, j}-\left[\mathbf{M}_{i j} \log \hat{\mathbf{M}}_{i j}+\left(1-\mathbf{M}_{i j}\right) \log \left(1-\hat{\mathbf{M}}_{i j}\right)\right].
\end{equation}

The total loss functions of the proposed framework are described as:
\begin{equation}
 \label{loss_total}
 \mathcal{L}_{\text {total}}=\mathcal{L}_{\mathrm{ce}}+\lambda_1 \mathcal{L}_{\mathrm{sim}} + \lambda_2 \mathcal{L}_{\mathrm{seg}},
\end{equation}
where $\lambda_1$ and $\lambda_2$ are the hyper-parameters used to balance these loss functions.
The parameters of network are updated via back-propagation.

\section{Experiments}
In this section, we first experimentally evaluate the effectiveness of the proposed algorithm against state-of-the-art techniques and investigate its robustness under unseen manipulation methods in Sec.\,\ref{result_discussion}.
Subsequently, we conduct an ablation study to explore the influence of proposed components in Sec.\,\ref{ablation_study}.
Finally, we demonstrate the interpretability of our approach through visualization analysis in Sec.\,\ref{visulization}.

\subsection{Experimental Setup}

\subsubsection{Datasets.} \label{dataset}
Following the convention, we adopt the widely-used benchmark dataset FaceForensics++ (FF++)~\cite{Rssler2019FaceForensicsLT} for training.
FF++ is a face forgery detection dataset consisting of 1000 original videos with real faces, in which 720 videos are used for training, 140 videos are reserved for validation and 140 videos for testing.
Then each video is forged by four state-of-the-art face manipulation methods, \emph{i.e.}, DeepFakes (DF)~\cite{Tora2018DF}, Face2Face (F2F)~\cite{Thies2019Face2FaceRF}, FaceSwap (FS)~\cite{Kowalski2018FS}, and NeuralTextures (NT)~\cite{Thies2019DeferredNR}.
Output videos are generated with different quality levels to create a realistic setting for manipulated videos, \emph{i.e.}, raw, high quality (HQ) and low quality (LQ), respectively.
To evaluate the robustness of our method, we also conduct experiments on the recent proposed large-scale face manipulated dataset, \emph{i.e.}, Deepfake Detection Challenge (DFDC)~\cite{Dolhansky2019TheDD}, Celeb-DF~\cite{Li2020CelebDFAL} and DeepfakeDetection\footnote{https://ai.googleblog.com/2019/09/contributing-data-to-deepfake- detection.html} (DFD).

\begin{table}[t]
\centering
\begin{tabular}{c|c|c|c|c|c}
\hline
\textbf{RGB} & \textbf{Freq} & \textbf{RFAM} & \textbf{MPSM} & \textbf{ACC} & \textbf{AUC} \\ \hline
$\surd$ & & & & 89.15\% & 93.03\% \\ \hline
$\surd$ & $\surd$ & & & 89.99\% & 93.13\% \\ \hline
$\surd$ & $\surd$ & $\surd$ & & 90.05\% & 93.78\% \\ \hline
$\surd$ & & & $\surd$ & 91.06\% & 94.96\% \\ \hline
$\surd$ & $\surd$ & $\surd$ & $\surd$ & \textbf{91.47\%} & \textbf{95.21\%} \\ \hline
\end{tabular}
\caption{Abalation study on the influence of different model components on FaceForensics++ (LQ) dataset.}
\label{ablatioin_study}
\end{table}

\subsubsection{Implementation.}
We implement the proposed framework via open-source PyTorch~\cite{Paszke2017AutomaticDI}.
For the frequency-aware cue, the $\alpha$ in Equ.\,\ref{freq} is empirically set to 0.33.
And the number of patches $k$ is set to 5.
To enhance the learning of local region relations, we set $\lambda_1$ and $\lambda_2$ in Equ.\,\ref{loss_total} to 10 and 1, respectively.
Following FaceForensics++~\cite{Rssler2019FaceForensicsLT}, we resize the input image to $299 \times 299$, and train the network using Adam optimizer~\cite{Kingma2015AdamAM} with a learning rate of 2e-4, a batch size of 32, betas of 0.9 and 0.999, and weight decay equal to 1e-5.
The total number of training epochs is set to 50, and the learning rate is reduced to half every 10 epochs.

\begin{figure*}[t]
	\centering
	\includegraphics[width=0.89\textwidth]{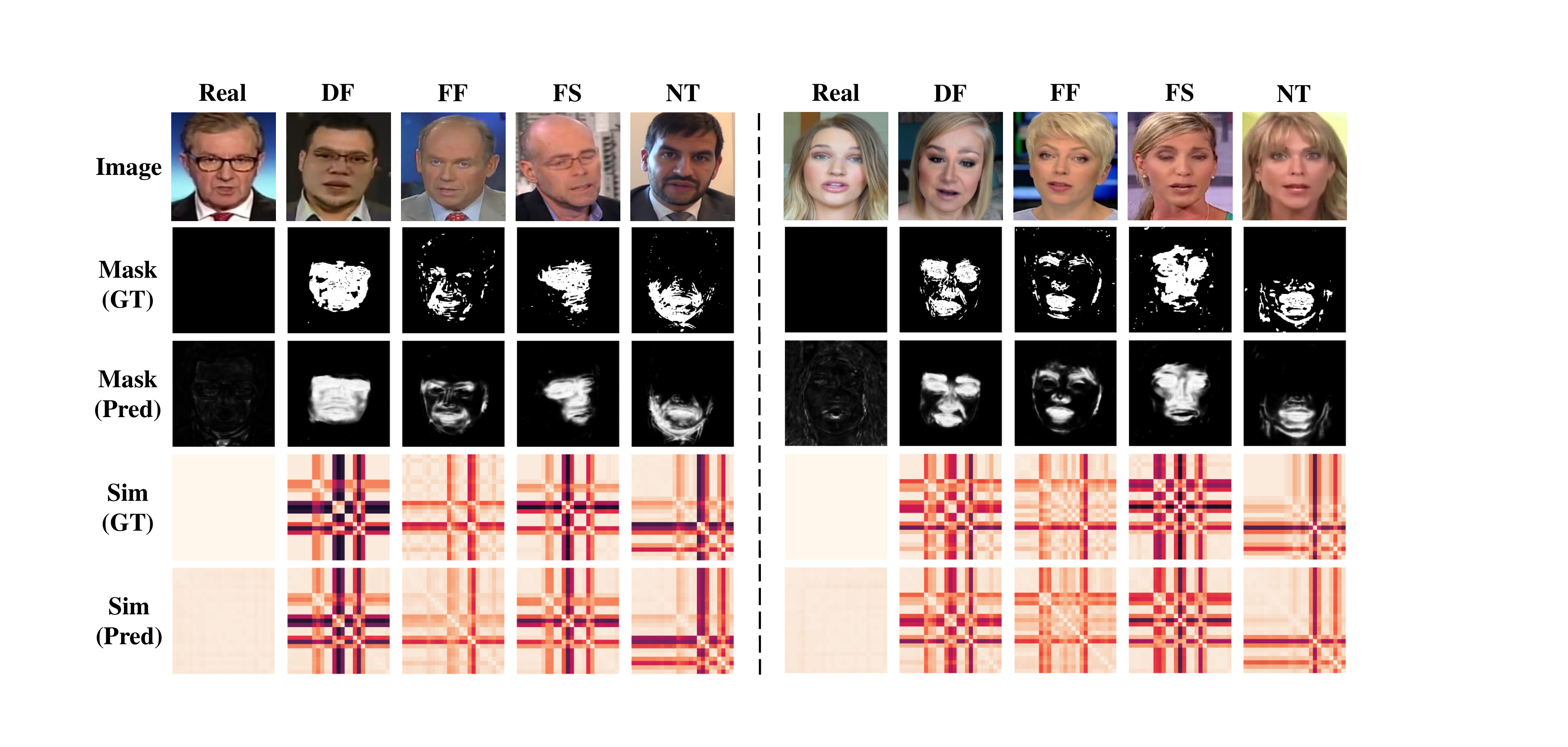}
	\caption{Visual examples of mask and local similarity patterns on various types of faces, \emph{i.e.}, Real, Deepfakes (DF), Face2Face (FF), FaceSwap (FS) and NeuralTextures (NT).}
	\label{visual_examples}
\end{figure*}

\begin{figure*}[t]
	\centering
	\subfigure[Real]{\includegraphics[width=0.327\textwidth]{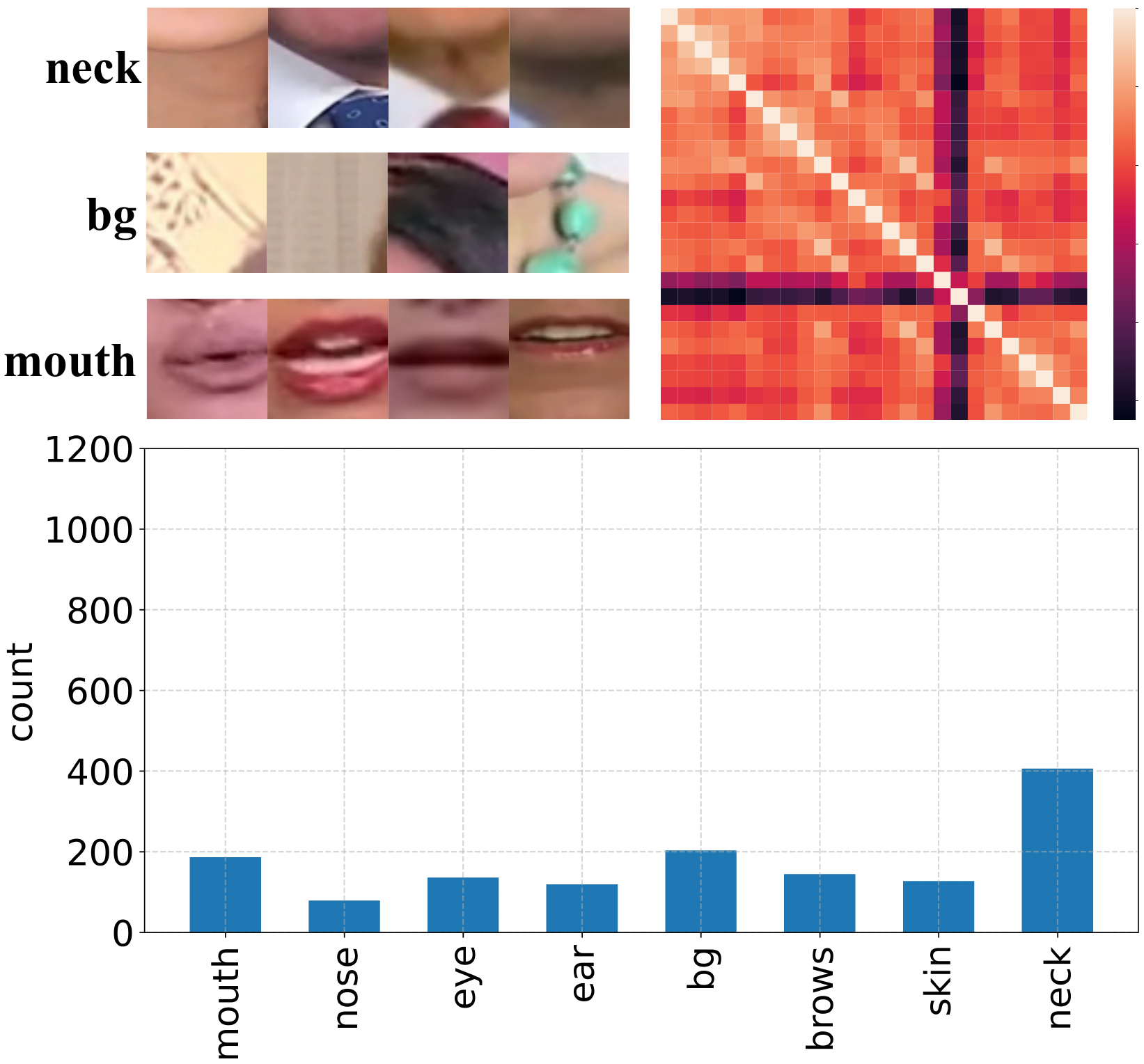}}
	\subfigure[Deepfakes]{\includegraphics[width=0.327\textwidth]{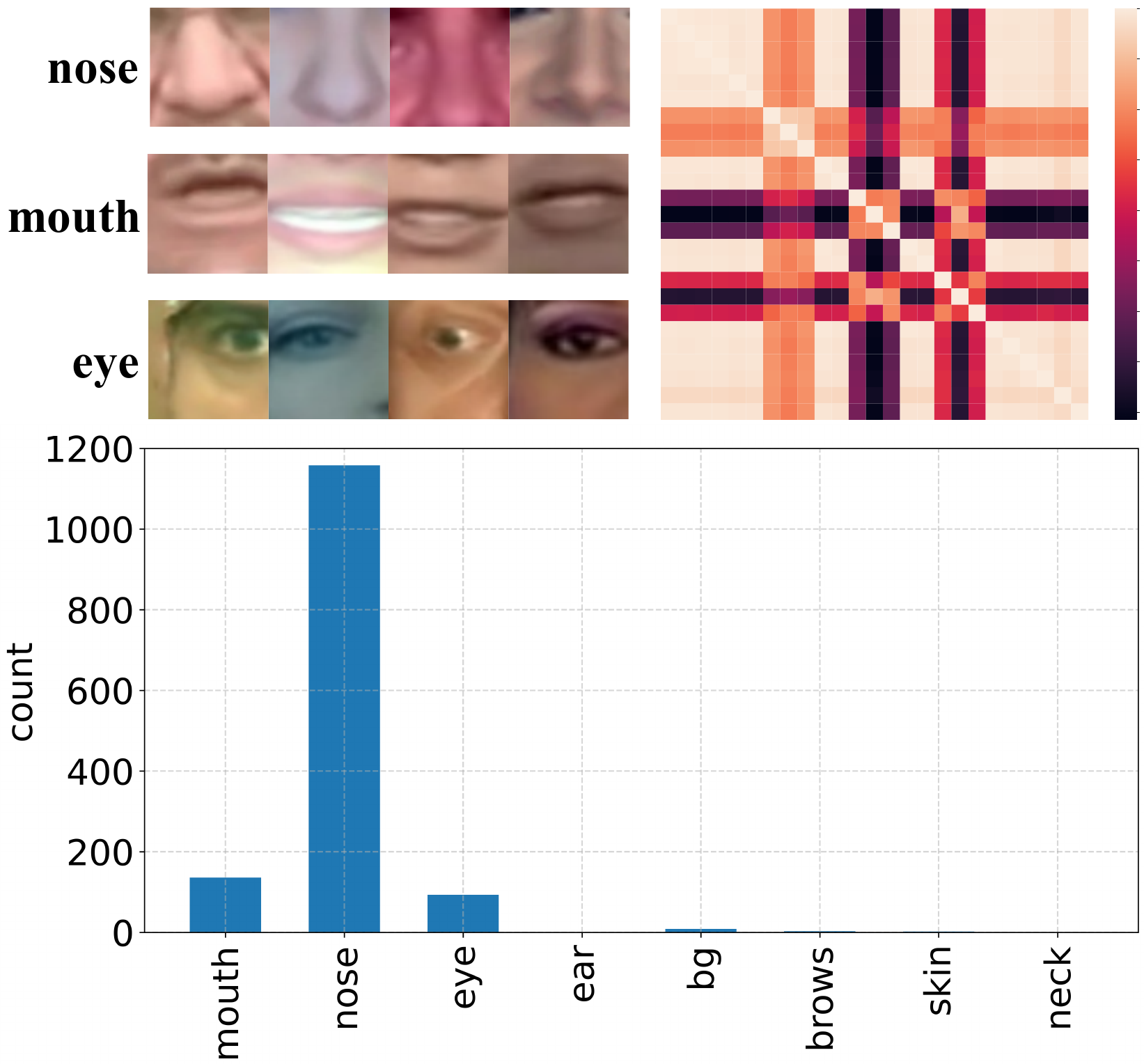}}
	\subfigure[NeuralTextures]{\includegraphics[width=0.327\textwidth]{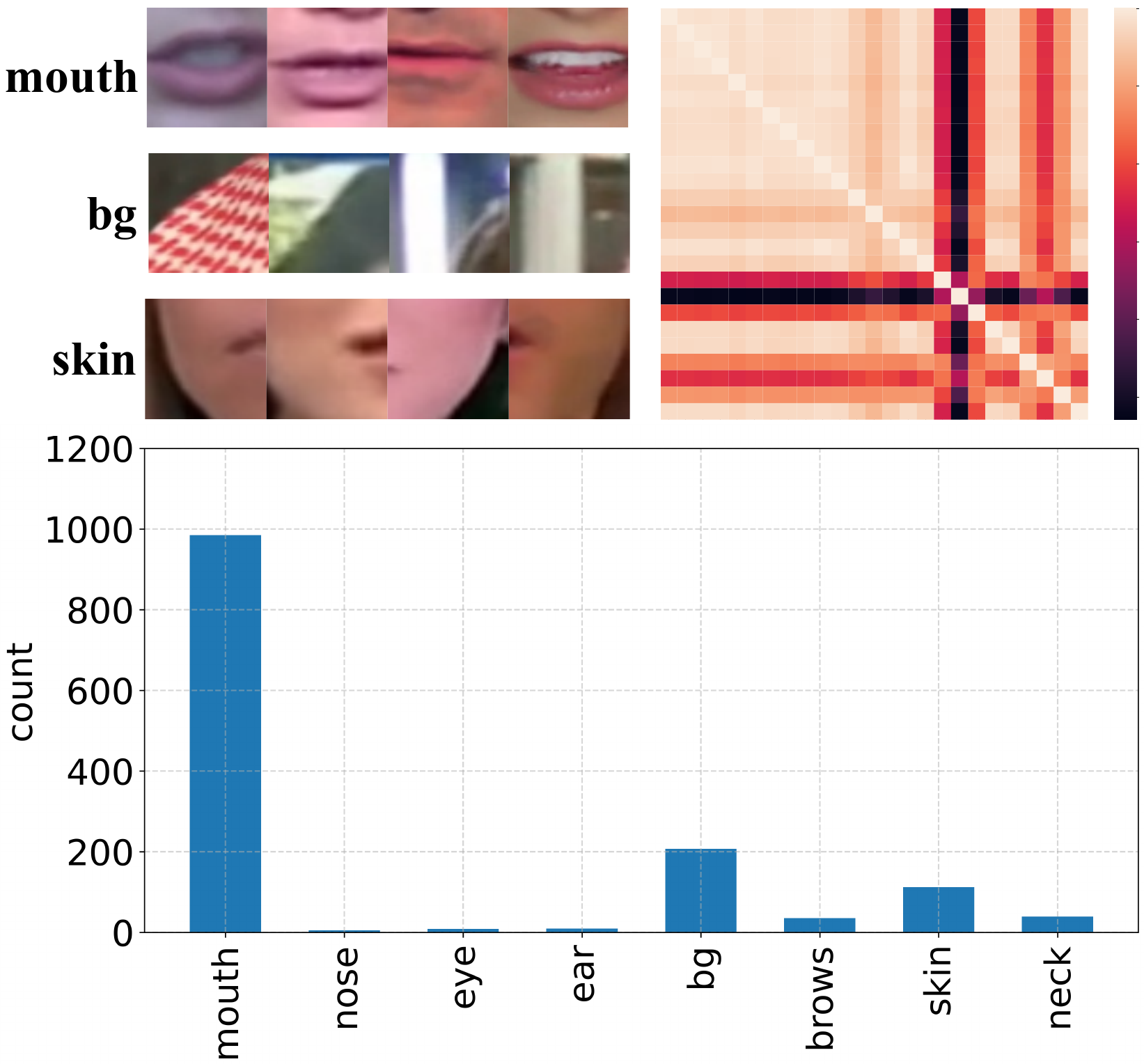}}
	\caption{The local similarity patterns and histograms of the most predictive patches on various types of faces.}
	\label{sim_patch_hist}
\end{figure*}

\begin{figure}[t]
	\centering
	\includegraphics[width=0.35\textwidth]{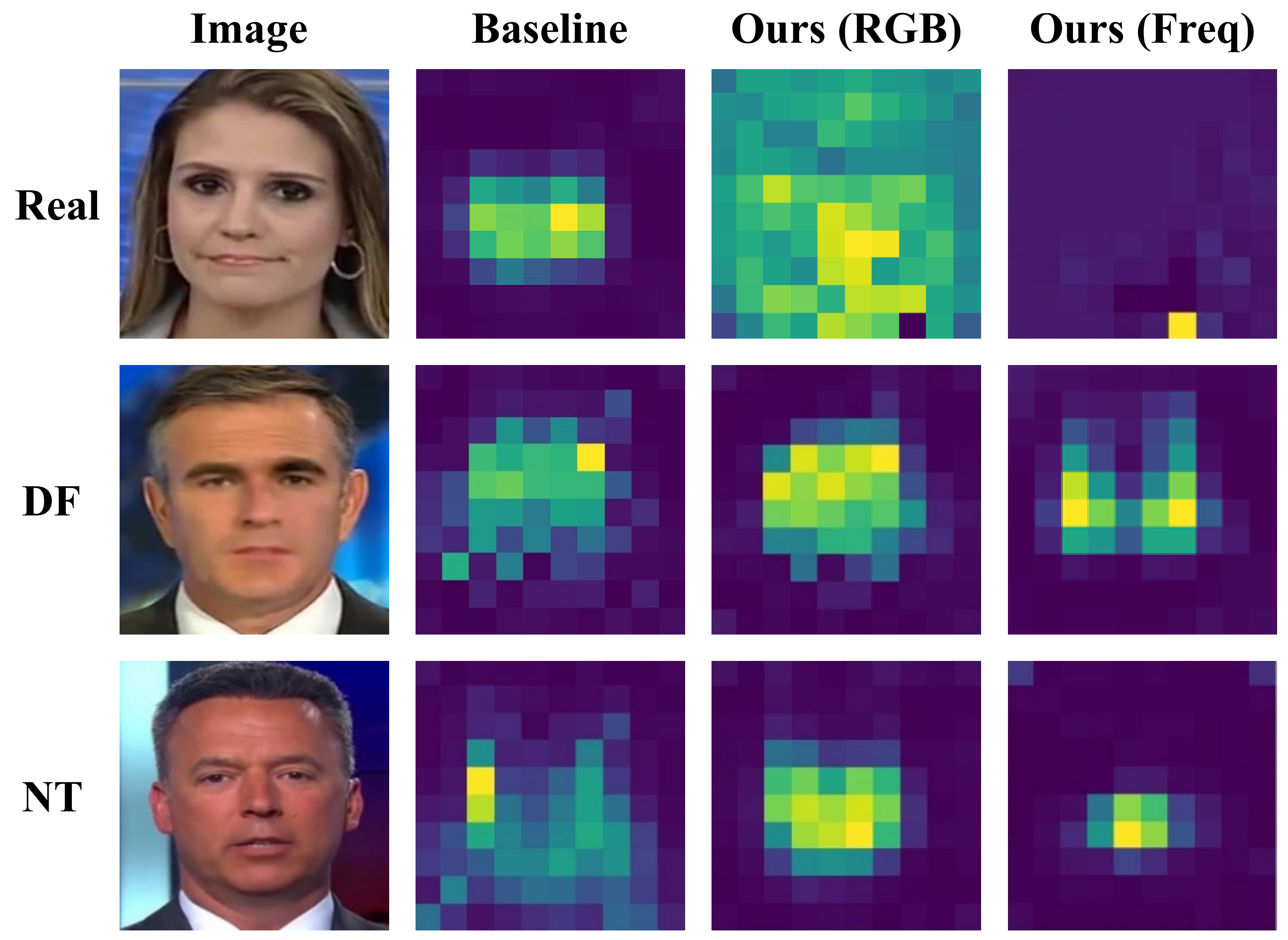}
	\caption{The feature maps of various types of faces extracted by baseline and our method, where the third and fourth columns are the feature map corresponding to RGB and frequecy, respectively.}
	\label{cam_rgb_freq}
\end{figure}

\begin{figure}[t]
	\centering
	\includegraphics[width=0.4\textwidth]{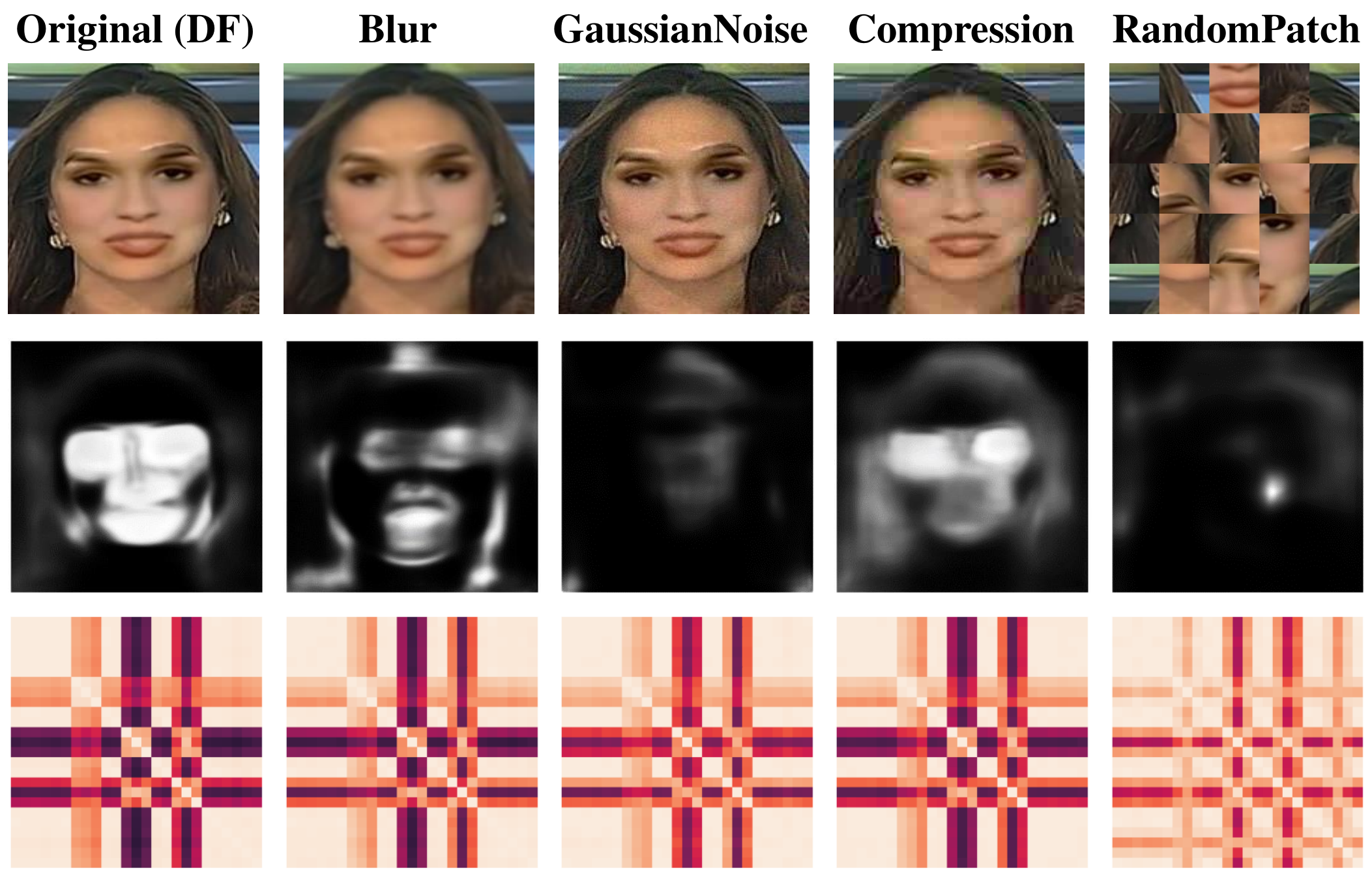}
	\caption{The visualization of the baseline (second row) and our method (third row) under different noises.}
	\label{visual_on_noises}
\end{figure}

\subsection{Results and Discussions} \label{result_discussion}
In this section, we apply the Accuracy score (ACC) and Area Under the Receiver Operating Characteristic Curve (AUC) as our evaluation metrics.
For a fair comparison, the results of all comparison methods are obtained from their paper.

We first evaluate the model performance under different quality settings (Raw, HQ and LQ) and the comparison with the state-of-the-art is showed in Tab.\,\ref{faceforensics}. 
We can observe that: 1) Our proposed method consistently outperforms all compared opponents by a considerable margin. 
For example, compared with the state-of-the-art F3 Net, the AUC of our method exceeds it by 0.12\%, 1.36\%, and 1.91\% at all the three quality settings, and this performance gain is also obtained under ACC.
Different from F3 Net which only utilizes the frequency information, our model combines both RGB and frequency information for a more comprehensively feature presentation, so that all kinds of artifacts of the forged face can be captured. 
2) Compared with Two-branch RN which also takes both RGB and frequency information into consideration, our method demonstrates superior ACC and AUC performance. This proves the effectiveness of our proposed RFAM.
3) Besides, our method has a significantly higher AUC on LQ than Face X-ray by 33.61\%.
To explain, Face X-ray relies on blending boundary that is susceptible to noise, whereas our method learns a content-independent local pattern through the proposed MPSM and is therefore more robust against disturbances including but not limited to image noise. This is important as the face forgery detection on low quality images is most challenging and measures the model's generalization ability to a large extent. 

Furthermore, we evaluate the proposed model on four different face manipulation methods listed in the FF++ dataset.
In the experiment setting, we follow the common practice by only using low-quality (LQ) videos in each manipulation method for training and test. 
The results are shown in Tab.\,\ref{single_test}.
It can be observed that our method consistently outperforms all comparisons on the four manipulation methods.
In particular, on the most challenging NeuralTextures (NT) manipulation method which does not produce noticeable forged artifacts, our model exceeds the state-of-the-art F3 Net by 5.99\% under the ACC metrics, which further illustrates the effectiveness of our proposed local relation learning.

To demonstrate the generalization capabilities of our proposed model, we perform an inter-test by training on the FF++ dataset while testing on Celeb-DF, DFDC and DFD datasets. Since Face X-ray introduces an additional dataset \emph{BI} in their experiment setting, we only include the results of Xception for a fair comparison. As shown in Tab.\,\ref{cross_dataset}, our method significantly outperforms Xception on all unseen datasets. The gain mainly benefits from that our method focuses on the intrinsic differences between real and forged regions, which are commonly present under all manipulated faces with various manipulation techniques. And our model carefully designs the similarity pattern in MPSM to describe this intrinsic difference which improves the model's generalizability to unknown scenarios.

\subsection{Ablation Study} \label{ablation_study}
To evaluate the effectiveness of the components of our proposed model, we develop the following variants: 1) the baseline model which contains only RGB input and is supervised by classification (Equ.\,\ref{loss_ce}) and segmentation (Equ.\,\ref{loss_seg}), 2) our method w/o RFAM and MPSM, which directly concatenate the final feature maps of two-stream network, 3) our method w/o MPSM, 4) our method w/o frequency and RFAM, where the input of MPSM is only RGB stream.

The quantitative results are listed in Tab.\,\ref{ablatioin_study}.
By comparing variant 1 and variant 2, the introduction of frequency information consistently improves the ACC and AUC.
The performance is further improved by fusing the two-stream information through the proposed RFAM.
It is worth noting that by using only MPSM, the ACC and AUC are substantially increased to 91.06\% and 94.96\%, which fully demonstrates the efficiency of the local relation features.
Combining all the proposed modules, our method achieves the best performance, 91.47\% and 95.21\% for ACC and AUC.

\subsection{Visualization} \label{visulization}

\subsubsection{Visual examples.}
Our framework makes predictions about the forgery based on the local similarity. 
The visual examples of mask and local similarity patterns on various types of faces are shown in Fig.\,\ref{visual_examples}.
It can be seen that the predicted similarity and mask are the same as the ground-truth, which well captures the local forged regions generated by different face manipulated algorithms.
For real faces, the features are similar between local regions, while forged faces show different similarity patterns depending on the specific manipulated regions.
Therefore, such discrepancy can be effectively served as a basis of classification.

\subsubsection{Interpretability of local relation.}
As shown in the Fig.\,\ref{sim_patch_hist}, we statistically estimate the similarity patterns, and histograms of the most predictive patches on various types of faces, \emph{i.e.}, Real, Deepfakes and NeuralTextures. %
Specifically, for each face type, we randomly sample 1000 images and average the similarity of local regions extracted by our method to obtain the corresponding similarity pattern (the top-right in Fig.\,\ref{sim_patch_hist}). We can observe that the similarity of real faces tend to be consistent across regions, while forged faces show specific patterns due to different manipulated regions, which indicates the local relations is a generalized pattern for the Face Forgery Detection task.
Besides, we select the most predictive region of each image based on the activation of the final feature map. For real face, the attention areas are evenly distributed, while fake face is mostly concentrated in forged regions such as the nose or mouth, which further proves the effectiveness and interpretability of our method.

\subsubsection{Complementarity of RGB and frequency information.}
Fig.\,\ref{cam_rgb_freq} presents the visualization of feature maps extracted by baseline and our method.
We can observe that:
1) The baseline has a higher response in the middle region of any type of face, while our method exhibits a different focus between real and forgery.
For example, in real faces, all regions of feature map have almost the same activation values, while in forged faces the activation values are larger only in the manipulated regions.
This is because our method imposes constraints on local features to ensure that the network learns a content-independent feature and focuses more on artifacts.
2) The frequency and RGB information of the forged faces are located in different regions, \emph{i.e.}, the former captures subtle forged boundary, while the latter shows higher response in facial areas such as eyes and mouth.
Through the proposed RFAM, the information can complement each other to further facilitate the learning of local features.
3) Finally, since the real face has not been manipulated, there is no anomalous response in the feature map of frequency stream.

\subsubsection{Robustness toward noises.}
Fig.\,\ref{visual_on_noises} shows the visualization of the baseline and our method under different noises.
Specifically, we apply different augmentations to the forged faces, \emph{i.e.}, blur, gaussian noise, compression, and random patch.
From Fig.\,\ref{visual_on_noises}, we can observe that the baseline has a significant error in the predicted masks under noises, especially under the gaussian noise and random patch.
This is because the baseline classifies based on global information and is sensitive to unseen patterns.
Conversely, the local similarity pattern of our method remains consistent under different noises.
Even if the face structure is broken, \emph{i.e.}, the random patch, the result is still robust.

\section{Conclusion}
In this paper, we introduce a new perspective for face forgery detection that models the relation of local regions. A novel architecture based on RGB-Frequency Attention Module is proposed with Multi-scale Patch Similarity supervision, which both considers RGB and frequency information and collaboratively learns comprehensive local relations. And the relations are further used for forgery detection. Extensive experiments and visualization demonstrate the robustness and generalizability of the proposed method on widely-used face forgery detection datasets.

\section*{Acknowledgements}
This work is supported by the National Science Fund for Distinguished Young (No.62025603), the National Natural Science Foundation of China (No.U1705262, No.62072386, No.62072387, No.62072389, No.62002305, No.61772443, No.61802324 and No.61702136) and Guangdong Basic and Applied Basic Research Foundation (No.2019B1515120049).

\bibliography{mybib.bib}
\end{document}